# Analogical Learning in Tactical Decision Games

## Tom Hinrichs, Greg Dunham, and Ken Forbus

Qualitative Reasoning Group
Northwestern University
1890 Maple Avenue
Evanston, IL 60201 USA
{t-hinrichs, gdunham, forbus}@northwestern.edu

### Abstract

Tactical Decision Games (TDGs) are military conflict scenarios presented both textually and graphically on a map. These scenarios provide a challenging domain for machine learning because they are open-ended, highly structured, and typically contain many details of varying relevance. We have developed a problem-solving component of an interactive companion system that proposes military tasks to solve TDG scenarios using a combination of analogical retrieval, mapping, and constraint propagation. We use this problem-solving component to explore analogical learning.

In this paper, we describe the problems encountered in learning for this domain, and the methods we have developed to address these, such as partition constraints on analogical mapping correspondences and the use of incremental remapping to improve robustness. We present the results of learning experiments that show improvement in performance through the simple accumulation of examples, despite a weak domain theory.

Content Areas: Case-Based Reasoning, Machine Learning

## Introduction

A longstanding challenge for machine learning is to learn from complex structured examples in broad, open domains. We believe that domain-independent analogical mapping and constraint propagation can form an effective foundation for such learning. Our experience applying these techniques to Tactical Decision Games led us to develop several strategies that make use of limited domain knowledge to assist in the transfer and adaptation of precedents. Although these additional techniques require some domain-specific knowledge, we believe them to be useful in a broad variety of domains.

We have been exploring analogical learning as part of developing interactive *companion systems* (Forbus and Hinrichs, 2004), software agents that learn over the long term. One important aspect of a companion is that it should learn from experience by accumulating examples. This is a weak form of learning that we expect to augment eventually with facilities for generalization, but it is a critical capability nevertheless. In this paper, we describe the problems, techniques, and experimental results of applying analogical learning to Tactical Decision Games.

## Tactical Decision Games

Tactical Decision Games are scenarios used by military personnel to hone their command skills (Schmitt, 1994). A scenario is provided in a page or two of text describing the situation, accompanied by a sketch map illustrating the terrain and what is known about force deployments. These scenarios are quite complex: Typically, several answers are equally reasonable, given different assumptions about what the enemy is doing. Because much of the critical information in a TDG is implicit in the spatial relations between units and terrain, they are a challenging vehicle for analogical learning and problem solving. To make the problem-solving task more tractable, we chose not to try to produce complete courses of action, but rather to suggest task assignments for Blue-Side military units that would each independently address the overall military objectives stated in the problem.

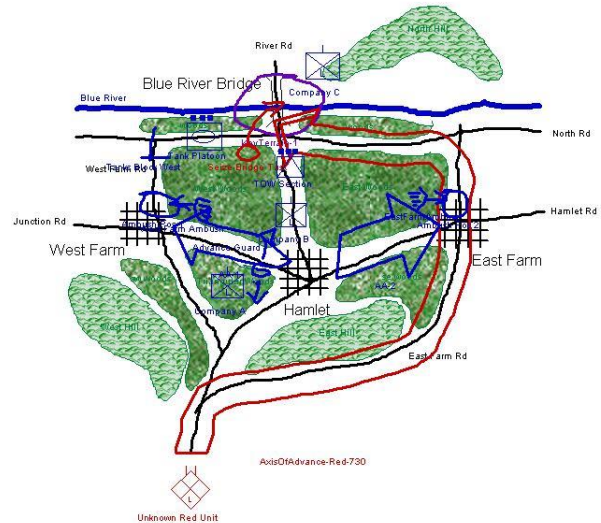

**Figure 1: A Typical TDG Sketch**

# Interactive Companion Systems

An interactive companion is intended to be a software assistant that helps you work through complex reasoning tasks in a domain by suggesting precedents, counter-examples and possible explanations. The companion system for Tactical Decision Games consists of three components: nuSketch Battlespace, an analogical tickler, and a facility for requesting task suggestions.

## nuSketch Battlespace

To encode TDG scenarios, we used nuSketch Battlespace (Forbus, Usher, and Chapman, 2003), a sketching application developed to facilitate strategic knowledge capture from military experts. The interface was designed using elements from US Army training manuals, and many of the representation and interface decisions were made in collaboration with military officers. The goal was to allow military commanders to describe scenarios (such as TDGs) in as natural a manner as possible. In addition to its role in knowledge acquisition, nuSketch Battlespace also provides facilities for spatial reasoning and extracts and translates spatial relations into propositional representations in a form suitable for inference and analogical mapping. The underlying knowledge base builds on the CYC ontology (Lenat, 1995), consisting of over one million facts. In addition to the sketching interface, nuSketch Battlespace (nSB) provides facilities for entering non-visual facts, such as commander's intent and enabling conditions and rationale for task selections.

## Analogical Tickler

As part of a companion system, retrieval of relevant precedents is conceived as an ongoing, automatic process. An "analogical tickler" runs as a distributed agent on a remote cluster computer in order to ensure that retrieval scales and does not become a bottleneck during an interactive session. The tickler uses MAC/FAC (Gentner and Forbus, 1995) to retrieve the precedent from the case library that is most similar to the user's sketch. As the user works, the analogical tickler receives continuous updates from the nuSketch application about the contents of the sketch. Once the previous retrieval has been completed, the tickler uses the updated information to re-run the retrieval and update its recommended precedent. While this does not guarantee that the precedent has been retrieved using the current state of the sketch as a probe, it is likely to have been retrieved using a fairly recent state of the sketch. The user is thus assured near-immediate access to a similar precedent at any point during the interactive session, even when using a very large case library during retrieval.

## Task Suggestions

In order to generate and present task suggestions, we integrated nuSketch Battlespace with the Companions architecture, and adapted its interface to include a "Suggestions" button in the toolbar. At any time while sketching a TDG-style scenario, the user may click the Suggestions button to request task suggestions for the sketch. Suggested tasks are generated via an analogy with the expert solution for the precedent, and presented visually on an overlay layer in the current sketch. A text summary describes facts about the suggestions that are difficult to represent visually, such as task types and targets.

Suggestions consist of proposals for one or more new tasks that may be assigned to units in the sketch. These tasks do not make up a unified plan, but are instead independent tasks that make sense based on the operational descriptions in the cited precedent. Although the generation of unified plans should ultimately be feasible, inferring coordinated military strategies from the current case base is beyond the scope of these experiments. Instead, we focus on learning lower-level tactics that can be directly transferred from one situation to another.

# Analogical Problem Solving for TDGs

Our system learns by accumulating examples and combining analogical mapping with constraint propagation to generate task suggestions based on a retrieved precedent. This is essentially a kind of case-based reasoning (Leake, 1996), although our emphasis is more on reasoning about and applying structural mappings. Our goal was not to build an expert system for TDGs, but rather to characterize analogical learning itself. In this section, we describe the core techniques enlisted in our analogical problem-solving experiments. In the following section, we describe several techniques used to improve upon the results obtained using this basic framework.

The process by which suggestions are generated consists of two distinct phases: retrieval and problem solving. The retrieval phase attempts to determine the best precedent from the case library, while the problem-solving phase attempts to apply the precedent's tactical solution to the user's sketch. Evaluating the phases separately is difficult because there is no domain theory that can be used to easily identify the 'best' precedent. Instead, a good precedent is characterized as one which can be used to generate good suggestions in the problem-solving phase.

## Retrieval

The first step in the analogical problem-solving process is to retrieve relevant precedents from a library of prior cases. Our library consists of 16 TDGs drawn from the book *Mastering Tactics* (Schmitt, 1994) and from the *Marine Corps Gazette*. The cases were encoded using NuSketch Battlespace and include a problem scenario and an expert solution to the problem. Both were sketched by the experimenters, who interpreted the graphical and textual representations found in the original sources. Identical terrain representations were used for both problem and solution. Fully encoded, a typical TDG problem or solution sketch contains between 500 and 1000 facts.

The retrieval process is invoked by opening a new problem sketch in NuSketch Battlespace and using the analogical tickler to return similar problems as precedents. We rely on the solution case to be sufficiently similar to the probe case that it can be mapped and applied to the new problem. We empirically tested this assumption as described in the experimental results section.

**Problem Solving**

The problem-solving phase begins by using SME (Falkenhainer, Forbus, and Gentner, 1989) to perform a structural comparison between the textbook solution to the retrieved precedent (the base of the analogy), and the user's current sketch (the target). The structural mapping introduces *analogy skolems*, or hypothesized entities that would exist if the cases were completely structurally aligned. The goal of the problem-solving phase is to fully resolve the skolemized Blue-Side tasks by using these structural mappings.

The primary method we use to resolve these skolemized task entities is constraint propagation. Note that the blue-side tasks in the precedent exist in a network of relations that capture the enabling conditions and rationale for the task. When these relations are mapped across to the target, they form candidate hypotheses that can serve as a constraint network. So, for example, if an ambush task is justified in the precedent because the attacking unit is close to the victim unit but not visible to them, then those criteria become constraints on the location of the task in the new situation.

We then create an anonymous instance of the task (called a *plunk*) which allows us to propagate constraints through this network. If the entities that constrain this task (such as its location) do not map directly to an entity in the target, a new plunk is introduced and its constraints are propagated again until concrete values are reached. In the next section, we describe how we enhanced this basic process with more specialized techniques to improve the quantity and reasonableness of information that could be transferred from a precedent.

## Improving Transfer Performance

In preliminary experiments, we found that the basic problem-solving approach sometimes yielded poor or even ridiculous results. Many of these failures arose from the fact that structure mapping is constrained only by the relations between entities, and not their types. For example, unconstrained structure mapping could result in a task proposal in which the West Woods is assigned to seize the distant city of Dullsville by following a path across the Blue River. We also found that it was never a good idea to plunk military units, which could usually be characterized as wishing the cavalry into existence. These examples makes painfully clear that the nature of the entities can be of vital importance when it comes to problem solving and transfer.

To avoid such problems and improve transfer performance, we developed several techniques for using limited domain knowledge to reason about skolemized entities and modify or reject candidate inferences for which there was insufficient domain support. A dependency tree was used to ensure well-founded support for plunks. We introduced *partition constraints* to prohibit mappings between fundamentally incompatible entity types. To relax structure mapping theory's requirement of 1-to-1 mappings between entities, we transferred from multiple alternative mappings. Finally, we applied limited reasoning and a weak domain theory to better assess the plausibility of transferred features.

**Calculating Plunk Dependencies**

Since our revised strategies for analogical problem solving abandon the assumption that all structurally-supported skolems can be successfully plunked in the solution, it becomes important to understand the relations between skolemized entities. If a task specification depends on the existence of an entity that may not get plunked, that task should not be plunked either since its description would be incomplete. We employed an algorithm to calculate plunk dependencies and drive the order in which they are examined for domain consistency. Once a plunk is rejected, all other skolems and candidate inferences that depend on its resolution will also be rejected.

**Partition Constraints**

Since structure-mapping theory considers only the structural relations between entities (Gentner, 1983), it's possible for a mapped entity to make no sense in the new context, despite a possibly deep structural similarity. To avoid such ontological mismatches, we added type partitions to the set of correspondence constraints accepted by the SME matcher. These *partition constraints* allow a user to specify that, for a given analogy, base entities from a particular collection in the ontology can correspond only to target entities of the same collection. For Tactical Decision Games, we required that terrain features map to terrain features, military units map to military units, and more specifically, that Blue Side units, tasks and paths map to Blue Side units, tasks and paths, and likewise for Red Side.

**Using Multiple Mappings**

Structure-mapping theory requires that a 1-to-1 correspondence exist between entities in a mapping. In cases where there are more entities in the base than there are in the target, the 1-to-1 constraint can lead to a large number of skolems. This can be problematic in the TDG domain because some types of entities just cannot be plunked: new units or terrain features cannot be forced into a TDG problem simply because they happen to fit an analogy.

The naïve solution is to discard these plunks as infeasible hypotheses. But in doing so, we are also forced to discard plunks and candidate inferences that depend on the initial plunk. A number of interesting features may not transfer, and the essence of a textbook solution may even be lost due to an unfortunate cardinality mismatch between problem and precedent.

Instead, we apply multiple structural mappings from the same base-target pair. To do this, we first impose a new correspondence constraint such that the previously unmapped base entity must correspond to some entity in the target. We then invoke SME to incrementally re-map. If a new consistent mapping cannot be found, the plunk is discarded. Otherwise, we modify the candidate inferences that referenced the original plunk to instead refer to the corresponding target entity in the new mapping.

By re-mapping, it is possible to introduce internal inconsistencies, such as a unit being in two places at the same time. Detecting this can require the use of some limited forms of domain reasoning, as described below.

## Using Limited Domain Knowledge

We found that by employing a weak domain theory, we were able to improve the hypotheses generated via the analogical problem-solving process. We used domain-specific reasoning to help resolve locations and paths, to disambiguate hypothesized entities, and to filter out infeasible suggestions.

### Spatial Reasoning

As a platform built to facilitate reasoning in the battlefield domain, nuSketch Battlespace offers a variety of powerful spatial reasoning capabilities. By using just a small subset of these features, we were able to markedly improve the suggestions offered to the user. Spatial reasoning was used to disambiguate or reject proposed paths and locations.

nSB's path planner was invoked for two purposes: to reject infeasible paths, and to disambiguate those that were feasible. A path suggested by analogy is described by specifying its starting location, end location, and (optionally) a set of visibility and/or trafficability constraints. Spatial reasoning facilities were used to plan the proposed path given the constraints. If a compatible path could not be planned, the path was rejected as an infeasible suggestion. If successful, the path was illustrated as an overlay to the user's sketch as part of the suggestions interface. Similar capabilities were used to evaluate suggested locations such as engagement areas or battle positions.

### Domain hierarchies

The nSB interface allows for articulation of the command hierarchy in a TDG. The hierarchy may consist of units that are physically present on the sketch, as well as organizational entities. A platoon in a scenario may be sketched, but their subordinate squads might be described using the non-visual facts editor. An analogy that suggests a path starting at the location occupied by one of the squads is problematic—where might we consider the squad to be located in the sketch? The problem is addressed using simple forward chaining to infer the squad's location to be the same as that of the commanding unit whose location is elaborated in the sketch.

## Experimental Results

In order to learn from examples, a system must be able to efficiently access good examples, transfer the relevant aspects of the precedent, and produce feasible and reasonable results. As more relevant examples are acquired, performance should improve. We performed two experiments to evaluate retrieval and problem-solving/learning performance. The first compares the actual retrieval performance to the similarity rankings produced by fully-constrained SME. The second experiment measures performance by systematically comparing generated task suggestions to tasks in the experts' solutions.

### Retrieval Experiments

The retrieval experiments were designed to determine how well the domain-independent MAC/FAC algorithm would scale to large cases and if retrieval accuracy would be swamped by irrelevant details in the cases, such as the specifics of where each terrain feature was as opposed to the overall tactical situation. We wanted to know if it was feasible to retrieve relevant cases without significant elaboration and preprocessing to characterize the tactical situation beforehand.

For each of 12 problem cases, we retrieved the 'best' precedents, according to MAC/FAC, from the library of 16 cases. (There was some overlap between problem cases and the library, for which we dynamically removed the problem case from the library.) Having retrieved the 12 best precedents, we then repeated the experiment with those cases removed from the library, and iterated this process five times in order to create a retrieval ranking of the top five precedents for each problem.

*How efficient is the analogical tickler? Does it scale?*

Throughout these trials, we found that the tickler and its MAC/FAC component preformed relatively well for the complex probes consisting of 500-1000 assertions. Once a sketch was loaded, the retrieval of relevant precedents was computed on the order of ten seconds or less, suggesting this is feasible for interactive behavior. Although it remains to be seen how this performance would hold up with a significantly larger library, we expect that it would only stress the computationally cheaper, feature-vector matching phase. Full structure mapping is only computed for the top three precedents that pass the MAC filtering stage.

*How accurate is the tickler?*

Because MAC/FAC is domain independent and pre-filters remindings based on surface-feature similarity, we wanted to see how well MAC/FAC retrieval corresponded to the structure-mapping similarity as measured by SME when provided with the correspondence constraints used by the problem solving process. For example, when we solve a tactical decision game from a prior case, we constrain the analogy mappings such that the Blue Side of the base maps to the Blue Side of the target. This is a bit of domain-specific knowledge that isn't available to the analogical tickler, so we would expect that sometimes the case retrieved from memory would not be the ideal precedent for problem solving. To assess the degree to which MAC/FAC limited performance, we constructed a complete similarity matrix. We then compared the retrieval rank with the unconstrained and constrained SME similarity rank.

We define retrieval accuracy as the rank difference between MAC/FAC and constrained SME, divided by the size of the library (which varies as we remove cases from it). So if MAC/FAC returns a precedent that should have been the second best precedent according to SME, then the error is 1/16, where the 16 is the size of the initial case library. For the five ranked retrieval tests over twelve problems, the average MAC/FAC error was 12% when compared to SME ranking without correspondence constraints, and 16% when correspondences are constrained. From this we can conclude that problems may not always be solved using the most structurally similar precedent when MAC/FAC is used for retrieval. The actual effect on problem solving was investigated in the problem-solving experiments described next.

## Problem-Solving Experiments

*How well does retrieval ranking correspond to the accuracy of proposed tasks?*
Rather than test learning directly by adding random cases to the library, we invoked the problem-solving process using the ranked precedents from the retrieval experiments. Then, to establish a long-term trend, we solved the problem with one of the three *least* similar precedents in a trial we referred to as "Evil MAC/FAC". Figure 2 shows average performance over 12 problems as a function of retrieval rank, where EM* denotes "Evil MAC/FAC". Presented this way, the chart shows how problem-solving performance corresponds to the similarity of the precedent as determined by MAC/FAC.

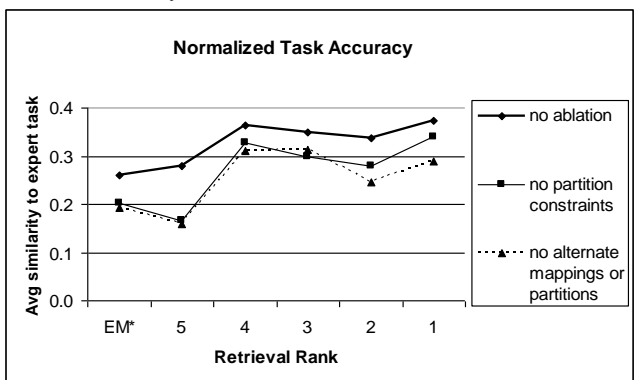

Figure 2: Problem-solving results

Because we didn't have a strong domain theory, we could not objectively and automatically evaluate complete tactical plans. Instead, we measured how similar the individual tasks produced were to some task in the expert solution to that problem. This provided a way to measure the reasonableness of tasks based on an expert's prior solution, without introducing the possible subjective bias of the experimenters. In some ways, this is an especially harsh criterion, because there can often be more than one solution to a problem. The point, however, was to show an improvement trend with the addition of cases, not to measure the absolute performance.

We assessed the quality of individual task assignments along four dimensions:

1) the type of task proposed
2) the target or object acted on
3) the unit assigned to perform the task, and
4) the location or path followed in the task.

Each generated task was compared to each task in the expert's solution along the four dimensions, giving partial credit for near misses. The accuracy was calculated as the highest similarity score, normalized by dividing by 25 (the highest possible score). So, for example, if the type of task was exactly the same (e.g., ambush), it received 5 points. If the proposed task was a specialization of the expert task, it received 4 points, the same tactical category (e.g., ambush vs. attack) received 3 points, a more general proposal got 2 points, the same posture (e.g., both offensive tasks or both defensive) received 1 point, otherwise it received nothing. Similar schemes were used for the other dimensions.

*How fragile is problem solving with respect to precedent?*
From Figure 2, we see that performance does not fall off sharply with retrieval rank. On the one hand, this suggests that the problem solving is robust and doesn't depend on working from just the right case. On the other hand, it also means that the learning curve is very shallow and performance improves only gradually. We would expect this in a task and domain for which there may be multiple satisficing solutions.

*What is the contribution of other problem-solving methods?*
In the ablation tests, we repeated the problem-solving trials first with the partition constraints turned off in the analogy mechanism, and again with both partition constraints and alternate mappings turned off. Here, we can see that these methods do make a noticeable difference in the accuracy of the solutions. Performance still improves overall, but the trend is less monotonic. The average performance is brought down in some trials when a problem is not solved at all, i.e., no tasks are proposed. However, without partition constraints, an even poorer precedent may still generate many proposed tasks which receive partial credit despite being infeasible.

## Discussion and Future Work

We were pleased to see a 30% improvement in the quality of tasks proposed from a poor precedent to a relatively good precedent. Given how structurally dissimilar the cases are in the corpus and the fact that the retrieval process uses no domain knowledge at all, the improvement suggests that the system can learn by accumulating examples without a strong domain theory.

Nevertheless, we were surprised by a number of problems. One reason why some cases were not solved well was that the retrieval process would sometimes return precedents that only made sense if the Blue Side in the problem were mapped to the Red Side in the solution. We could address this either by modifying MAC/FAC to accept correspondence constraints or by changing the problem-solving process to try solving with crossed-allegiance mappings.

A more general lesson is that although SME drives mapping based on relational structure, the nature of the entities still matters when it comes to transfer and problem solving. We partitioned correspondence constraints to filter out infeasible mappings and used multiple mappings to relax isomorphism constraints. Given an analogy skolem, we used limited domain knowledge to distinguish entities that could be transferred, such as tasks and engagement areas, from those that could not, such as mountains. This kind of domain knowledge can be viewed as *determinations* (Davies and Russell, 1987), rules that govern projectability in a given context.

Our approach to focusing on relevant features is somewhat different than that taken in most systems for inductive concept learning (cf. Blum and Langley, 1997). Rather than characterize relevance over a set of exemplars, we use structural criteria exclusively to guide retrieval. In the problem-solving phase, we use the constraint network contained in the structural mapping to select relevant features, and then use limited domain reasoning to filter inconsistent or infeasible transfers. This reduces our dependence on having a representative sample of cases, which is important for broad, open domains.

There are a number of ways to improve our problem-solving process. A stronger domain theory could yield more feasible and reasonable suggestions, if not unified, coherent plans. One possibility would be to generalize the tactical pattern in the precedent solution, for example, to "apply combined arms" or "find a gap in the enemy forces". With sufficient domain expertise, one could transfer the abstract pattern and re-operationalize it in the new situation.

From our perspective, however, a more important goal is to have more interactive and instructable operation. For example, we might like our system to learn partition constraints for a new domain, even if it is through explicit instruction from a user. Consequently, we will continue to explore these issues in the Companions project, although we intend to move on from Tactical Decision Games to interactive strategy games in order to support more interactivity and provide opportunities for reflective learning and user modeling.

## Acknowledgements

This research was supported by the Defense Advanced Research Agency.